# Attention-based Pairwise Multi-Perspective Convolutional Neural Network for Answer Selection in Question Answering

Jamshid Mozafari[a], Mohammad Ali Nematbakhsh[a], Afsaneh Fatemi[*,a]

[a] *Faculty of Computer Engineering, University of Isfahan, Isfahan, Iran*

**Abstract**

Over the past few years, question answering and information retrieval systems have become widely used. These systems attempt to find the answer of the asked questions from raw text sources. A component of these systems is Answer Selection which selects the most relevant from candidate answers. Syntactic similarities were mostly used to compute the similarity, but in recent works, deep neural networks have been used, making a significant improvement in this field. In this research, a model is proposed to select the most relevant answers to the factoid question from the candidate answers. The proposed model ranks the candidate answers in terms of semantic and syntactic similarity to the question, using convolutional neural networks. In this research, Attention mechanism and Sparse feature vector use the context-sensitive interactions between questions and answer sentence. Wide convolution increases the importance of the interrogative word. Pairwise ranking is used to learn differentiable representations to distinguish positive and negative answers. Our model indicates strong performance on the TrecQA Raw beating previous state-of-the-art systems by 1.4% in MAP and 1.1% in MRR while using the benefits of no additional syntactic parsers and external tools. The results show that using context-sensitive interactions between question and answer sentences can help to find the correct answer more accurately.

**Keywords:** Question Answering system, Factoid question, Deep learning, Convolutional neural network, Attention mechanism, Pairwise ranking

## 1. Introduction

Human need for information about a particular subject is called the Information Need. The answer to this need can be text, audio, image, video, or a combination of them (Kolomiyets & Moens, 2011). One of the essential information needs that humankind has been involved with throughout history is to answer the questions that have emerged in the minds of humankind over and over again. Nowadays, web space has built up a vast repository of data with extremely high redundancy (Brill, Dumais, & Banko, 2002). Thus, it can be an excellent resource to find answers. That's why, many users try to find pages related to their information needs and questions on the Internet and going through them to extract the answer.

There are several ways to find answers using the Internet, such as forums, social networks, search engines, and question answering systems. QA systems return specific parts of the document information as an answer. This answer may be a word, a sentence, a paragraph, or an audio/video clip (Dwivedi & Singh, 2013). There are two kinds of question answering systems containing Knowledge-based and Information Retrieval-based (IR-based) systems. In knowledge-based systems, documents have a regular structure, or they are converted into a regular structured such as relational databases or knowledge databases. In relational databases and knowledge databases, objects and their attributes are stored at the database, and the communication between them is defined. However, IR-based systems usually handle queries and unstructured documents. Unstructured information has more detailed information than structured information. That's why nowadays more attention has been relatively drawn on this type of information which is used more than knowledge-based systems. IR-based systems have three general phases containing Question Processing (QP), Phrase Retrieval (PR), and Answer Processing (AP) (Jurafsky & Martin, 2014). The overall structure of an IR-based system is shown in Figure 1.

---

[*] **Corresponding Author.**
  *Email addresses:* mozafari.jamshid@eng.ui.ac.ir *(J. Mozafari),* nematbakhsh@eng.ui.ac.ir *(M. A. Nematbakhsh),* a_fatemi@eng.ui.ac.ir *(A. Fatemi).*

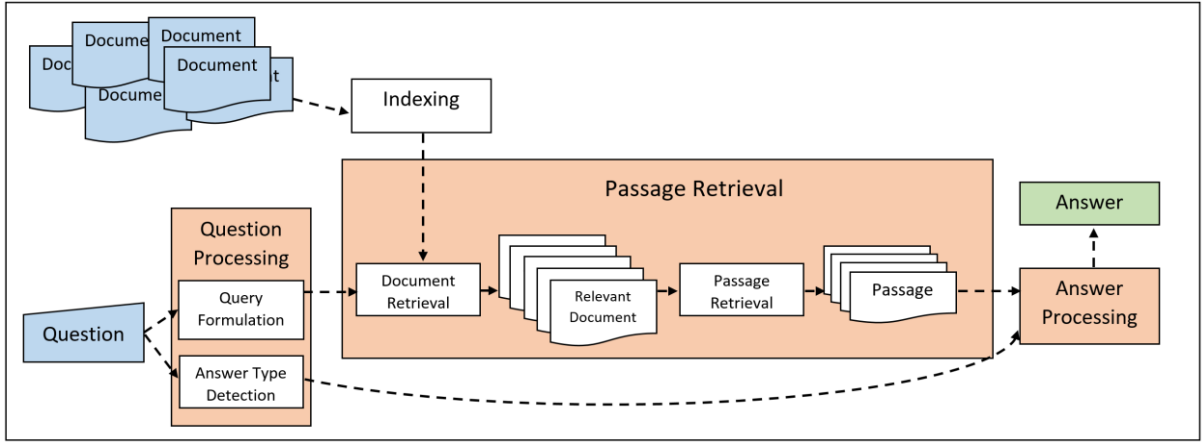

**Figure 1: The structure of an IR-based Question answering system (Jurafsky & Martin, 2014)**

In the QP phase, the user's question is processed, and the question information is extracted. This phase includes Query formulation and the Answer type detection sections. Query formulation creates a query from the user's question keywords and sends it to the PR phase. Answer type detection attempts to discover the answer type using various analyzes on the words and structure of the user's question. PR phase retrieves related phrases from the documents after receiving the query. It seems that the "document" is not a suitable unit for QA systems. It is necessary to examine different parts of the document, such as paragraphs and sentences. After analyzing different parts of the document, this phase sorts the phrases by the relevance and presents them. In the AP phase, the question's answer is extracted from the phrases returned from the PR phase. This phase presents question answers in two ways: the extracted answer, and the generated answer. The extracted answer extracts the question's answer from the text, but the generated answer generates answers based on different rules using the answers found in the text (Jurafsky & Martin, 2014; Mishra & Jain, 2016).

Determining the semantic similarity between two sentences is one of the essential issues in the field of natural language processing and information retrieval. This issue, used in paraphrase recognition, is also called sentence matching(Magnolini, 2014). Another application of this method is to find the most semantically relevant answer to a question in the AP phase of the QA systems. This method is very suitable for factoid questions because the sentence structure of the answer of the factoid question is very similar to the meaning and structure of the question. In the QA systems where the AP phase operates based on the answer extraction, there is a particular component called the Answer Selection (AS). This component is responsible for identifying the best answer to the question as the correct one from candidate answers. The AS can be considered as a supervised learning problem. Given $q=\{q_1, q_2, \ldots, q_n\}$ where $q_i$ refers to a question, each $q_i$ comes together with a candidate answer set of $\{(s_{i1}, y_{i1}), (s_{i2}, y_{i2}), \ldots, (s_{im}, y_{im})\}$, in which $s_{ij}$ refers to the $j^{th}$ candidate answer, and $y_{ij}$ refers to the answer label. If $j^{th}$ candidate answer is the correct one to $i^{th}$ question, then $y_{ij}=1$, and vice versa. A classifier can be trained to predict candidate answers for unseen questions with this labelled data (Yih, Chang, Meek, & Pastusiak, 2013). Indeed, the classifier predicts the semantic similarity between question and answer (Echihabi & Marcu, 2003). Classifier's input is $(q_i, a_{ij})$ that each refers to a sentence, and the output is $y_{ij}$, which is 0/1 that represents the correctness of the answer. For example, consider the q question in Table I, with three candidate answers $a_1$ and $a_2$ and $a_3$. This classifier must output zero for inputs $(q, a_1)$ and $(q, a_2)$, and output one for input $(q, a_3)$.

**Table I: An example from TRECQA Raw (Wang, Smith, & Mitamura, 2007)**

| Q | Where is the highest point in Japan? | |
|---|---|---|
| $a_1$ | Climbing Mt Fuji is more than just getting to the top of a mountain. | $y_{11} = 0$ |
| $a_2$ | No camping is allowed on Mt Fuji. | $y_{12} = 0$ |
| $a_3$ | It was not alone, just one of the hundreds gathering on the summit of Mt Fuji, at 12,388ft the highest point in Japan. | $y_{13} = 1$ |

Various ways to solve the answer selection problem has been presented so far. These methods can be divided into two general categories. The first category includes the methods which try to measure the similarity between question and answer using feature engineering and syntactic methods. The second category includes methods which use deep neural networks. Over the past few years, neural network and deep learning have become more prominent and have used to solve the answer selection problem. In this paper, we use deep neural networks and try to provide a model with a convolutional neural network (CNN). The proposed model does not use external sources, such as WordNet (Miller, 1998), syntactic parsers (Jurafsky & Martin, 2014) and named entity recognizer (NER) (Jurafsky & Martin, 2014). Instead, the model uses the raw texts of the question and the answer.

The contribution of this research paper includes:

- We proposed the Attention-based Pairwise Multi-Perspective Convolutional Neural Network that ranks the candidate answers in terms of semantic and syntactic similarity, using convolutional neural networks.
- The attention mechanism and a sparse feature vector are used to capture context-sensitive interactions between a question and an answer.
- Wide convolution is used instead of the narrow convolution to increase the importance of the interrogative word.
- The MAP and MRR measures of the proposed model shows that it performs better than the state of the art.

In section II, related works will be explained., The proposed model will be described in detail in section III. In section IV, the proposed model will be evaluated with TrecQA Raw (Wang, et al., 2007) dataset, and the accuracy will be compared to the other models. Finally, the paper will be concluded in section IV.

## 2. Related Works

The early works have used string overlap features to measure sentence similarity. These have included features such as bag-of-words overlap and grams overlap (Wan, Dras, Dale, & Paris, 2006). However, these approaches cannot capture linguistic and semantic features. For example, the question "Who did Jamshid call?" cannot be answered with the "Ali called Jamshid," although these two sentences have the most overlap. Hence, the bag-of-words features are poor for sentence similarity (Surdeanu, Ciaramita, & Zaragoza, 2008). To solve this problem, various approaches have used lexical resources like WordNet to consider synonymous words (Fernando & Stevenson, 2008). In these approaches, it was difficult to reconcile with words that were not mentioned in lexical sources.

Many approaches have used semantic and syntactic structure to capture the linguistic and semantic features. In question answering, it is possible to use dependency tree for the question and the candidate answers. The answers can be ranked based on increasing order of edit distance between the question dependency tree and each answer dependency tree (Punyakanok, Roth, & Yih, 2004).

In recent years, new methods have been developed to solve the answer selection problem using neural networks and deep learning. Neural network methods aim to eliminate the need for manual feature adjustment, but they need a lot of input data to extract the features, which as a constraint is not a simple challenge to overcome.

Yu et al. (Yu, Hermann, Blunsom, & Pulman, 2014) have introduced the first work for AS using neural networks. The proposed model uses two methods to represent inputs. In the first method, bag-of-words is used that computes the sum of the sentences words' vectors and normalizes them. In the second method, the bigram is used in such a way that the word vectors are considered as bigram using convolutional neural networks. This model focuses on the use of the convolutional neural network in answer selection and does not consider any semantic relationships between question and answer. Feng et al. (Feng, Xiang, Glass, Wang, & Zhou, 2015) have combined convolutional neural networks with fully-connected neural networks in different ways and have produced various models. These models come from the combination of hidden layers, convolutional operators, pooling operators, and the activation function. In these models, the ranking method is changed to the pairwise which shows that using pairwise ranking is better than pointwise ranking. Severyn et al. (Severyn & Moschitti, 2015), instead of using various models, have provided a lightweight model which has shorter training time than other models. This model is the first one which uses feature vectors and wide convolution and shows that using external vectors can be useful. Tay et al. (Tay, Phan, Tuan, & Hui, 2017) have presented a model similar to Severyn et al. (Severyn & Moschitti, 2015), but they have used recurrent neural networks instead of convolutional neural networks. This model shows recurrent neural networks have a better understanding of the context and use external vectors as well. He et al. (He, Gimpel, & Lin, 2015) have presented a model using a convolutional neural network. This model uses a multi-perspective convolutional neural network instead of using a lightweight model such as Severyn et al. (Severyn & Moschitti, 2015) and several different models such as Feng et al. (Feng, et al., 2015). The model does not use any external components to demonstrate that the neural network alone is sufficient for answer selection. Tu (Tu, 2018) has shown that some components of MPCNN model (He, et al., 2015) can be omitted without altering the accuracy of the model. The model has also shown that the attention mechanism can be useful to some extent. Rao et al. (Rao, et al., 2019) have presented the HCAN model that bridges the gap between relevance matching and semantic matching. They believed that the model captures the rich presentation of sentences. Rao et al. (Rao, He, & Lin, 2016) have presented a pairwise model which have used MPCNN (He, et al., 2015) as a pointwise model. This model shows pairwise is better than pointwise for answer selection. Shen et al. (Shen, et al., 2018) developed the KABLSTM model, which utilizes knowledge graphs. They developed a context-knowledge interactive learning architecture, which used interactive information from input sentences and knowledge graph. The related works are shown in Table II.

Table II: Related works with their characteristics.

| Reference | Network | Attention Mechanism | Wide Conv. | Sparse Vector | Pairwise Ranking | MAP | MRR |
|---|---|---|---|---|---|---|---|
| (Yu, et al., 2014) | CNN | ✗ | ✗ | ✗ | ✗ | 0.711 | 0.785 |
| (Feng, et al., 2015) | CNN | ✗ | ✗ | ✗ | ✓ | 0.711 | 0.800 |
| (Severyn & Moschitti, 2015) | CNN | ✗ | ✓ | ✓ | ✗ | 0.746 | 0.808 |
| (Tay, et al., 2017) | RNN | ✗ | ✗ | ✓ | ✗ | 0.750 | 0.815 |
| (He, et al., 2015) | CNN | ✗ | ✗ | ✗ | ✗ | 0.762 | 0.830 |
| (Tu, 2018) | CNN | ✓ | ✗ | ✗ | ✗ | 0.762 | 0.830 |
| (Rao, et al., 2019) | CNN, RNN | ✓ | ✓ | ✗ | ✗ | 0.774 | 0.843 |
| (Rao, et al., 2016) | CNN | ✗ | ✗ | ✗ | ✓ | 0.780 | 0.834 |
| (Shen, et al., 2018) | CNN, RNN | ✓ | ✗ | ✗ | ✗ | 0.792 | 0.844 |

## 3. Model Architecture

In this paper, we propose an Attention-based Pairwise Multi-Perspective Convolutional Neural Network model, AP-MPCNN. This model follows the Siamese structure (Bromley, et al., 1993) in which there are two parallel subnets, each of which processes an input sentence. All parameters are shared in these two subnets. Figure 2 shows the AP-MPCNN architecture.

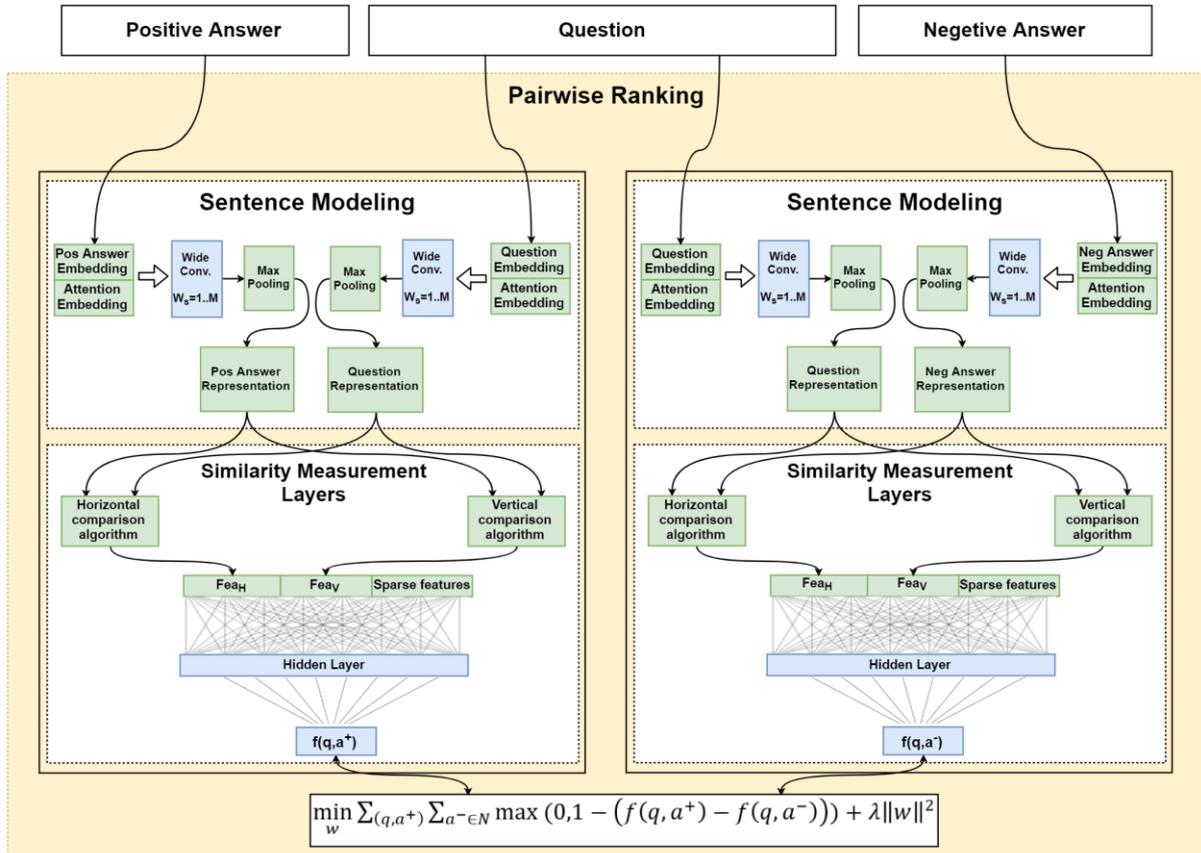

Figure 2: AP-MPCNN architecture. The blue components are trainable and the green components are non-trainable.

AP-MPCNN is a model which uses a convolutional neural network. This model contains Sentence Modeling and Similarity Measurement Layers. The sentence modeling produces a vector using convolutional neural networks, which captures the meaning of two input sentences. The similarity measurement also measures the similarity between input sentences using a fully connected neural network and the similarity vector. More precisely, in this model, the convolutional operator of variable window size is applied to the question sentence and the answer sentence embeddings (He, et al., 2015). The convolutional operator produces convolutional feature maps. Then, for each one, the max-pooling operator is applied separately and for each sentence, various vectors are represented which produce a matrix. Then, two different algorithms for measuring the similarity between the question and the answer are applied to each row and column corresponding to the question and the answer matrices (Tu, 2018). Afterward, by sending the final vector to a fully connected network, the correctness of the candidate answer is measured. Finally, the calculated similarity for the positive answer-question and the negative answer-question are ranked using the pairwise ranking. In the following, each section of the model presented in Figure 2 will be explained.

## 3.1. Sentence Modeling

In the AP-MPCNN model, each sentence is first entered into the model as a text string. But text sentences are not suitable to use in deep neural networks and should be converted to number vectors. In this research, pre-trained Glove (Pennington, Socher, & Manning, 2014) is used. In the Glove, for every word, there is a 300-dimensional vector which captures the word semantic. To present each sentence, first, the sentence should be tokenized. If there is a word in the sentence that is not in the Glove, <unk> would be used instead of the word. Then, word is replaced with the vector extracted from the Glove. Instead of <unk>, a vector initialized with uniform distribution U[-0.25: 0.25] is used. Consequently, each sentence is converted to a matrix $R^{d \times |s|}$ in which |s| is the length of the input sentence and d is the length of the word vector. For example, Figure 3 refers to a matrix $R^{300 \times 4}$ that shows the embedding matrix for "Who is Jamshid?".

| Who | is | Jamshid | ? |
|---|---|---|---|
| -0.2456 | 1.4896 | 0.1362 | 0.9863 |
| 0.1345 | -0.0012 | -0.1496 | -1.1456 |
| ... | ... | ... | ... |
| 1.1598 | 0.7321 | 0.2314 | 0.0036 |
| 0.2496 | 1.9637 | 0.0021 | -0.3276 |

Figure 3: Embedding matrix of "Who is Jamshid?"

This action is performed for both input sentences. In this research, one of the sentences is the question, and another is one of the candidate answers.

### 3.1.1 Attention

The Attention mechanism (Bahdanau, Cho, & Bengio, 2015) allows models to focus on the parts of the input that are more related to the output word, for each output word. This mechanism is commonly used in recurrent neural networks, especially neural machine translation models. Thus, to be used in convolutional neural networks, the attention mechanism must be adapted to the convolutional neural network.

In sequence to sequence models (Sutskever, Vinyals, & Le, 2014), because output words are generated separately, the attention calculation is done separately for each word, and the output word is generated. But in the AP-MPCNN model, both sentences are given as inputs to the model. As a result, at first, the attention mechanism can be applied to both inputs; weighted vectors are given as inputs to the model so that the model focuses on essential words for each output (He, Wieting, Gimpel, Rao, & Lin, 2016). In the following, applying attention mechanism to the question and the candidate answer will be explained:

1- If $S_1 \in R^{d \times |q|}$ (q refers to the question sentence) is the question and $S_2 \in R^{d \times |a|}$ (a refers to the answer sentence) is the candidate answer, the attention matrix $D \in R^{|q| \times |a|}$ is produced which each element determines the similarity between a question word and an answer word.

$$D[i][j] = \cos(S_1[:][i], S_2[:][j]) \tag{1}$$

2- For each input word, the attention weight is calculated; that is, the elements of each row are summed and $E_1 \in R^{|q|}$ is produced. This action is also performed for each column separately, and $E_2 \in R^{|a|}$ is produced.

$$E_1[i] = \sum_j D[i][j] \tag{2}$$

$$E_2[j] = \sum_i D[i][j] \tag{3}$$

3- $E_1$ and $E_2$ are converted into normalized vectors to determine the effect of each word on another sentence, that is, the softmax function (Buduma & Locascio, 2017) is applied to both $E_1$ and $E_2$, and vectors $A_1$ and $A_2$ are generated.

$$softmax(z)[i] = \frac{e^{z[i]}}{\sum_j e^{z[j]}} \; ; for\; i = 1,2,\dots,|z| \tag{4}$$

$$A_1 = softmax(E_1) \tag{5}$$

$$A_2 = softmax(E_2) \tag{6}$$

4- Each element of $A_1$ and $A_2$ is multiplied by the corresponding element in the vectors $S_1$ and $S_2$, and the effect of the words is changed according to their importance in answering the question.

$$newEmb_1[:][i] = A_1[i] \times S_1[:][i] \tag{7}$$

$$newEmb_2[:][i] = A_2[i] \times S_2[:][i] \tag{8}$$

5- $newEmb_1$ and $newEmb_2$ are concatenated with the $S_1$ and $S_2$ and new embedding matrices are produced known as the Attention embeddings.

$$attenEmb_1 = concat(S_1, newEmb_1) \tag{9}$$

$$attenEmb_2 = concat(S_2, newEmb_2) \tag{10}$$

This first specifies the word's semantic and then the model will understand the context-sensitive interactions between the question and the answer more specifically. Produced attention embedded will be sent as input to the convolutional neural network.

*3.1.2 Convolution Layer*

The AP-MPCNN model uses holistic convolution (He, et al., 2015). In holistic convolution, during a convolutional operation, all dimensions of word vector are covered. Holistic convolution can be known as "temporal" convolution since 1D convolution is performed over the sequence of tokens in the order they are visited or over time. The reason why this type of convolution and each filter are called holistic convolution and a holistic filter respectively is that each filter spans all dimensions of the word vector. Most of the models presented in related works used narrow convolution. If the sentence has a length |S| and the window size is w, then the narrow convolution will produce a convolutional feature map∈R$^{|S|-w+1}$ (w≤|S|). In narrow convolution, some words are not sufficiently involved. One of the most important words in the question is the interrogative word used at the beginning of the sentences. In narrow convolution, less attention is paid to the words at the beginning and the end of the sentence containing the interrogative words. As a result, the interrogative word, which is one of the main components for finding the correct answer in the sentence, will have the least effect. Accordingly, the use of wide holistic convolution (Kalchbrenner, Grefenstette, & Blunsom, 2014) can be useful and will further increase the value of interrogative words. In wide holistic convolution, the zero margins are added to cover equally all the sentence words. That's why the AP-MPCNN model uses wide holistic convolution. Figure 4 shows the structure of a wide holistic convolution.

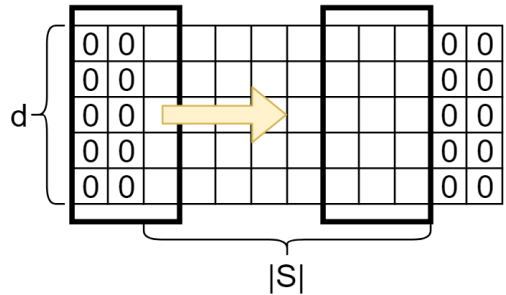

**Figure 4: Wide holistic convolution sliding on the embedded**

Each F filter is represented by a tuple $<w_s, W_F, b_F, h_F>$. In this tuple $w_s$ is the sliding window length, $W_F \in R^{d \times ws}$ is the filter parameters, $b_F$ is the value bias and $h_F$ is the activation function. By applying wide holistic convolution, a vector $out_F \in R^{|S|}$ is generated which the element i is the equivalent

$$out_F(w_s, S)[i] = h_F(W_F . S[:][i: i + w_s - 1] + b_F) \quad (11)$$

Where S[:][i:i+ws-1] denotes the i$^{th}$ to (i+ws-1)$^{th}$ columns of S inclusive and $1 \leq i \leq |s|\text{-ws+1}$.

### 3.1.3. Pooling

AP-MPCNN uses Max pooling to extract the most important features from convolutional feature maps. Tu (Tu, 2018) defines some objects composed of groups and blocks used to classify convolutions. A group is a tuple consisting of a convolutional layer with window size, a pooling function, and an embedded sentence. A group is defined as equation (12) shows:

$$group(w_s, pooling, sent) = pooling(out_F(w_s, sent)) \quad (12)$$

A block also refers to a set of groups as shown below. In this research, pooling refers to as the max pooling.

$$block = \{group(w_s, max, sent): w_s \in \{1 \ldots M\}\} \quad (13)$$

### 3.1.4. Sliding window sizes

AP-MPCNN uses different sliding window sizes to test different grams. This idea has been used by Kim (Kim, 2014) to classify sentences using convolutional neural networks. Figure 5 shows a subnet from the input to the final pooling stage.

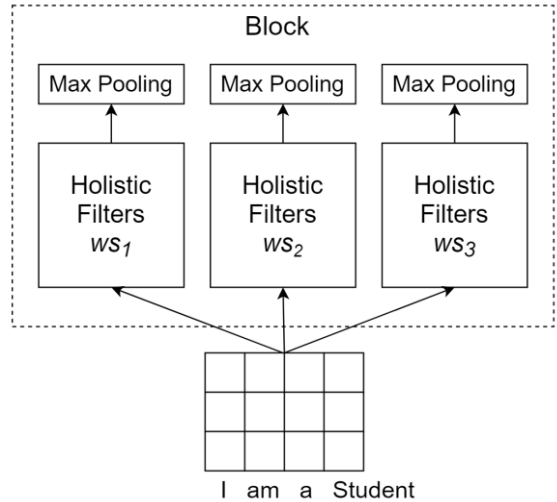

**Figure 5: Converting input to convolutional feature maps**

### 3.2 Similarity Measurement Layer

For this layer, both of the horizontal and the vertical comparison algorithms are used. These algorithms use comparison units that include similarity metrics which consist of the cosine distance, the Euclidean distance, and the element-wise absolute differences. The first type of comparison unit, comU$_1$, uses all three types of comparison units, while the second type of comparison, comU$_2$, uses only the cosine distance and the Euclidean distance. These choices are based on experience. These two comparison units are shown below.

$$comU_1(x, y) = \{\cos(x, y), L_2 Euclid(x, y), |x - y|\} \quad (14)$$

$$comU_2(x, y) = \{\cos(x, y), L_2 Euclid(x, y)\} \quad (15)$$

He et al. (He, et al., 2015) use two comparison algorithms to measure the similarity of the sentences. These algorithms are containing Horizontal comparison algorithm and Vertical comparison algorithm. In the horizontal comparison algorithm which is shown in Figure 6, regM is a |Filters|×M matrix and regM$_i$[*][ws$_j$]∈R$^{|Filters|}$ is a feature vector which group(ws$_j$, max, sent$_i$) produces. This algorithm produces a vector called fea$_H$∈R$^{2\times|Filters|}$ which captures the similarity of each row of regM$_1$ and regM$_2$ using comparison units which comU$_2$ includes.

```
HORIZONTAL − COMPARISON(S_1, S_2)
   fea_H = []
   for width w_s = 1 … n, ∞
      regM_1[*][ws_1] = group(w_s, max, S_1)
      regM_2[*][ws_1] = group(w_s, max, S_2)
   for i = 1 … numFilters
      fea_h = comU_2(regM_1[i], regM_2[i])
      fea_H.extend(fea_h)
return fea_H
```

**Figure 6: Horizontal comparison algorithm which captures the similarity of each row (He, et al., 2015)**

The vertical comparison algorithm, shown in Figure 7, uses wide holistic convolution for variable window sizes and produces a vector called $fea_V \in \mathbb{R}^{3 \times M^2}$ which captures the similarity of each one using comparison units which comU$_1$ includes.

```
VERTICAL − COMPARISON(S_1, S_2)
   fea_V = []
   for width w_{s_1} = 1 … n, ∞
      oG_1 = group(w_{s_1}, max, S_1)
      for width w_{s_2} = 1 … n, ∞
         oG_2 = group(w_{s_2}, max, S_2)
         fea_v = comU_1(oG_1, oG_2)
         fea_V.extend(fea_v)
return fea_V
```

**Figure 7: Vertical comparison algorithm which captures the similarity of each column (He, et al., 2015)**

### 3.2.1 Sparse feature vector

Sequera et al. (Sequiera, et al., 2017) propose a manual feature vector to measure the similarity of the two sentences, which increases the model accuracy and can be useful in the neural network. Due to the tiny size of the vector in comparison with the final vector of the neural network, this vector is called Sparse Feature Vector, consisting of four elements. In the following equations, SW refers to stop words, Q refers to the question, and A refers to the answer.

- The ratio of the number of shared words between two sentences to the total number of words in two sentences.

$$\frac{|Q \cap A|}{|Q|+|A|} \quad (16)$$

- The ratio of the number of shared words between two sentences words except the stop words to the total number of words in two sentences.

$$\frac{|Q-\{SW\} \cap A-\{SW\}|}{|Q-\{SW\}|+|A-\{SW\}|} \quad (17)$$

- The ratio of the weighted sum of shared words between two sentences to the total number of the words of two sentences.

$$\frac{\sum idf_{Q \cap A}}{|Q|+|A|} \quad (18)$$

- The ratio of the weighted sum of shared words between two sentences except the stop words to the total number of words of two sentences.

$$\frac{\sum idf_{Q-\{SW\} \cap A-\{SW\}}}{|Q-\{SW\}|+|A-\{SW\}|} \quad (19)$$

In these equations, the word weight is equivalent to the Inverse Document Frequency (IDF) measure (Manning, Raghavan, & Schütze, 2008). In this way, the weight of each word is based on word frequency in the text body. The goal is to show the importance of the word in the text. This is used in information retrieval and data mining. The word weight is improved with increasing word repetition in the text, but if it appears in more documents, it is probably a common word and does not have much value. This weight is calculated as shown in equation (20). Here, N specifies the total number of the documents in the corpus, and $n_w$ specifies the number of documents which w is used in.

$$idf(w) = \log \frac{N}{n_w} \tag{20}$$

Figure 8 shows the sparse feature vector.

| $\frac{|Q \cap A|}{|Q| + |A|}$ | $\frac{|Q - \{SW\} \cap A - \{SW\}|}{|Q - \{SW\}| + |A - \{SW\}|}$ | $\frac{\sum idf_{Q \cap A}}{|Q| + |A|}$ | $\frac{\sum idf_{Q-\{SW\} \cap A-\{SW\}}}{|Q - \{SW\}| + |A - \{SW\}|}$ |
|---|---|---|---|

**Figure 8: Sparse feature vector**

In questions about named entities, the entity names in the question are also shown in the correct answer. So, using this vector and concatenating it to the final feature vector can identify better answers.

The Horizontal-Comparison algorithm produces a one-dimensional vector of size 2×|Filters| vector and the Vertical-Comparison algorithm produces a one-dimensional vector of size 3×$M^2$ vector. By concatenating the sparse feature vector of size 4 with these two vectors, $H_v$ will be generated.

$$|H_v| = 2 \times |Filters| + 3 \times M^2 + 4 \tag{21}$$

This vector captures information about each pair of the candidate answer and the question. Figure 9 shows the final feature vector $H_v$.

| fea$_H$ | fea$_V$ | Sparse Vector |
|---|---|---|

**Figure 9: Final feature vector ($H_v$)**

### 3.2.2 Fully Connected Layer

The $H_v$ is considered as the input to a fully-connected network for measuring similarity. In equation (22), $W_{h1}$ refers to training parameters of the fully connected layer and $b_{h1}$ refers to bias. The intermediate layer is produced as:

$$H_L = \tanh(W_{h1} H_v + b_{h1}) \tag{22}$$

In the final layer, the probability of candidate answer correctness is measured. In equation (23), $W_{h2}$ refers to training parameters of the fully connected layer and $b_{h2}$ refers to bias. The final layer is produced as:

$$f(q,p) = sigmoid(W_{h2} H_L + b_{h2}) \tag{23}$$

Because of the high density of parameters, Dropout algorithm is used to improve regularization. Dropout prevents feature co-adaptation by setting to zero a portion of hidden units during the forward phase (Srivastava, Hinton, Krizhevsky, Sutskever, & Salakhutdinov, 2014). The structure of the shallow neural network is shown in Figure 10.

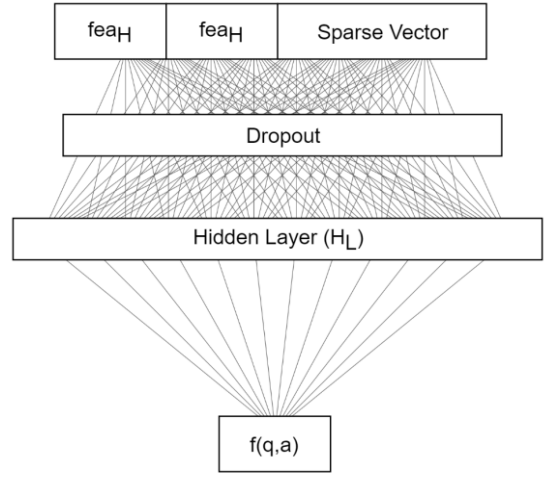

**Figure 10: Shallow neural network for similarity measurement**

*3.3 Pairwise Ranking*

Rao et al. (Rao, et al., 2016) develop a pairwise model which uses pointwise models as base components. The model presented so far can be used as a pointwise model in the pairwise model. The proposed model can be considered as a function f(q,p), which returns a probability for answer correctness. The pairwise model attempts to minimize the difference between the positive answer and the question and maximize the difference between the question and the negative answer using triplet loss function. In below, the triplet loss function is shown:

$$TripletLoss(q, p^+, p^-, \lambda, w) = \min_w \sum_{(q,p^+)} \sum_{p^- \in N} \max(0, 1 - (f(q, p^+) - f(q, p^-))) + \lambda \|w\|^2 \quad (24)$$

In equation (24), q refers to the question, $p^+$ refers to the positive candidate answer, $p^-$ shows the negative candidate answer, N is the set of negative candidate answers for q, w is the model parameters and λ shows the regularization parameter.

The pairwise model considers N in two modes. In the first mode, the set members are randomly selected from the negative candidate answers. In the other mode, the set members are selected from the negative candidate answers which are most similar to the positive candidate answers. In this research, the second mode is chosen because it causes the model output to be smaller for negative candidate answers and the model works better.

The pairwise ranking is expected to return a more massive similarity score for the positive answer and a smaller score for the negative answer. Triplet ranking loss function which is shown in equation (24) is evaluated to learn the joint representation of answers.

## 4. Experiments and Results

For answer selection in QA systems, the MAP and MRR metrics are used that are also proposed in information retrieval systems for the ranked list (Manning, et al., 2008). Both metrics measure the quality of the ranked list. In this research, the trec-eval tool[*] developed by the TREC community is used to measure these metrics.

MAP considers the average precision over different levels of recall. In equation (25), Q is the set of questions, $m_j$ is the number of relevant answers for the $j^{th}$ question, $R_{jk}$ is a ranked list containing the top k answers from most relevant to least relevant answer of the $j^{th}$ question determined by the model, and Precision(R) is the ratio of relevant answers to the total number of answers in the R.

$$MAP(Q) = \frac{1}{|Q|} \sum_{j=1}^{|Q|} \frac{1}{m_j} \sum_{k=1}^{m_j} Precision(R_{jk}) \quad (25)$$

MRR is considered only the rank of the top result. $r_j$ is the rank of the first actual relevant answer in the rank list for the $j^{th}$ question.

$$MRR(Q) = \frac{1}{|Q|} \sum_{j=1}^{|Q|} r_j \quad (26)$$

---

[*] **https://github.com/usnistgov/trec_eval**

## 4.1 Dataset

The TrecQA Raw dataset is prepared by Wang et al. (Wang, et al., 2007) which is from the TREC Question Answering tracks 8-13. Tracks 8-12 data are used for training and data from track 13 is used for validation and testing. Candidate answers in the validation and testing set are selected from sentences in the question's document pool. These questions contain at least one non-stop word that is also in the question and humans judge their relevance. The average length of sentences in TrecQA Raw is 18. Statistics about the TrecQA Raw dataset is summarized in Table III.

**Table III: Statistics of the TrecQA Raw dataset**

| Split | Number of Questions | Number of Sentence Pairs | Percent of Relevant Pairs |
| --- | --- | --- | --- |
| Train | 1227 | 54317 | 13.6% |
| Validation | 81 | 1148 | 19.3% |
| Test | 95 | 1517 | 23.0% |
| Total | 1403 | 56982 | 14.1% |

## 4.2 Detail of Implementation

We implement the proposed model with PyTorch library (Subramanian, 2018) in Python programming language and train the model on NVIDIA GeForce GTX 1050. TorchText library is used to process input sentences. We do not use a particular library to tokenize input sentences; instead, the split function of string class in Python is used. The batch size is equal to 32. We consider a vector initialized with uniform distribution U[-0.25: 0.25]. Glove word vectors initialized with 64B-300d (Pennington, et al., 2014) are used for word embeddings. The number of holistic filters of the convolutional neural network is equal to 300. Thee window size is tuned in the range from 1 to 3. The Max Pooling is applied for pooling operation. The number of hidden units is equal to 300. The dropout is set to 0.5. We use Tanh function for activation function.

To train the proposed model, we set the learning rate to 0.00018. The model is trained for 6 epochs. The learning rate patience is set to 2, and the learning rate reduce factor is equal to 0.3. It means that after every 2 epoch, the learning rate is multiplied by 0.3. The number of negative samples is equal to 8, and the regularization is set to 0.0006405. The optimizer algorithm is Adam.

As shown in Figure 2, some components are trainable and some of them are not. The total number of training parameters is about 1270k. The number of parameters in the convolution layer is 300×(1+2+3)×600. The number of parameters in the fully connected layer is 300×(2×300+3×9+4)+300. Hence the total number of training parameters is 1080k+189k≈1270k.

## 4.3 Ablation Study

We present an ablation study of the proposed model, comparing the original model with 4 ablation baselines. In each ablation baseline, we just use one of the components containing Attention, Wide convolution, Sparse feature vector, and Pairwise ranking. Figure 11 shows the MAP and MRR of each ablation baseline, respectively.

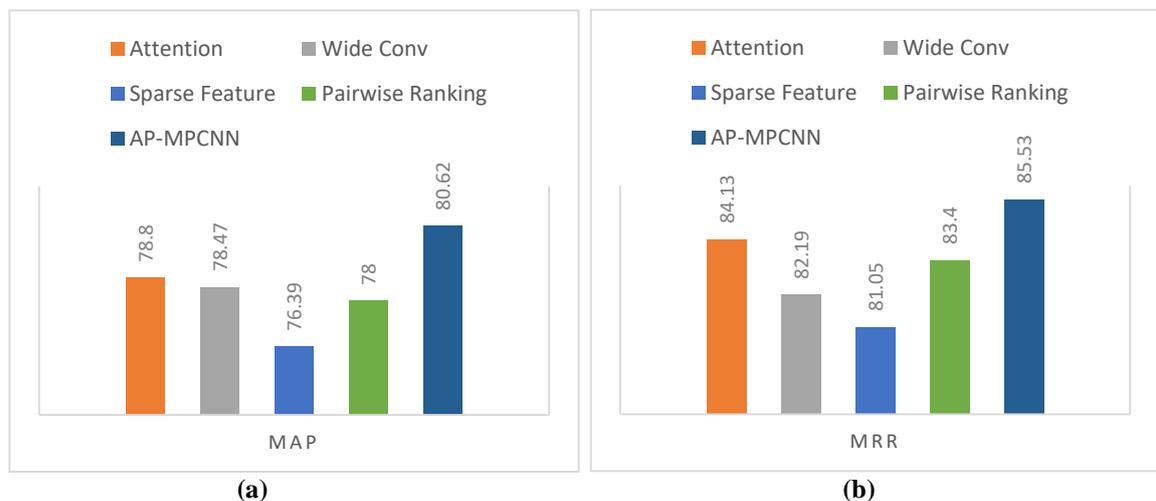

Figure 11: MAP score and MRR score of each ablation baseline

We also test the proposed model in 16 different situations. The results are shown in Table IV.

**Table IV: MAP score and MRR score for different situations**

|     | Epoch | | Batch Size | | Window Size | | # of Holistic Filter | | # of Hidden Units | | # of Negative Samples | | Regularization | | Learning Rate | |
| --- | --- | --- | --- | --- | --- | --- | --- | --- | --- | --- | --- | --- | --- | --- | --- | --- |
|     | 5 | 7 | 16 | 64 | 2 | 4 | 200 | 500 | 150 | 450 | 5 | 10 | 0.0001 | 0.001 | 0.0001 | 0.001 |
| MAP | 78.6 | 80.6 | 80.0 | 78.0 | 77.6 | 79.4 | 79.5 | 77.9 | 78.4 | 77.6 | 79.7 | 79.4 | 78.3 | 77.9 | 76.7 | 76.0 |
| MRR | 82.9 | 85.5 | 84.4 | 82.5 | 81.4 | 84.6 | 84.3 | 82.5 | 84.8 | 83.6 | 83.7 | 83.4 | 83.1 | 81.9 | 82.2 | 82.4 |

### 4.4 Results

The state-of-the-art of answer selection models, reviewed in related works, and the AP-MPCNN are compared in Table V, according to the MAP and MAP.

**Table V: Comparison of proposed model and other answer selection models, according to MAP and MRR**

| Reference | MAP | MRR |
| --- | --- | --- |
| (Yu, et al., 2014) | 0.711 | 0.785 |
| (Feng, et al., 2015) | 0.711 | 0.800 |
| (Tay, et al., 2017) | 0.750 | 0.815 |
| (He & Lin, 2016) | 0.758 | 0.822 |
| (He, et al., 2015) | 0.762 | 0.830 |
| (Tu, 2018) | 0.762 | 0.830 |
| (Tay, Tuan, & Hui, 2018) | 0.770 | 0.825 |
| (Rao, et al., 2019) | 0.774 | 0.843 |
| (Rao, et al., 2016) | 0.780 | 0.834 |
| (Shen, et al., 2018) | 0.792 | 0.844 |
| **AP-MPCNN** | **0.806** | **0.855** |

The results presented in Figure 11 and Table IV, show that each of the components containing Attention mechanism, Wide convolution, Sparse feature vector and Pairwise ranking, is not enough to improve an answer selection model and all of them should be used. An important note in AP-MPCNN model is that these components are not add training parameters to the model and just try to increase the comprehension of question and answer sentences. For example, the Attention component creates word vectors without needing training parameters. Other components also do not need training parameters and just increase model accuracy with matrix calculation.

### 5. Conclusion

In this paper, four ideas were used to improve accuracy. The reasons for using each idea were described separately and their effects have been reviewed. These ideas can be divided into two groups:the first group consists of pairwise ranking, and the second group consists of attention mechanism, wide holistic convolution, and sparse feature vector.

The first group makes a better ranking. This ranking method is very suitable for models with two inputs. The AP-MPCNN also has two different inputs for the candidate answer.

The second group makes it possible to use all the contents of the sentences and the context-sensitive interactions between questions and answer sentences. In models following the Siamese structure, two inputs are examined individually and are only related to the similarity measurement layer. This feature can limit the model and eliminates the useful information that can be obtained from the relationship between the two sentences. Therefore, using the relationship between two inputs can improve model accuracy.

As mentioned above, we can say that the first group is useful for solving the answer selection problem, and can be applied to any model. But the second group shows that the accuracy of the model can be increased by using all the contents of the sentences and the context-sensitive interactions between question and answer sentences.

Due to the widespread use of the attention mechanism, a better and faster mechanism is offered using this mechanism which is more robust than the attention mechanism. This mechanism is the Transformer (Vaswani, et al., 2017), which uses parallel attention mechanism. Using this mechanism and external tools can also improve the accuracy of the model.

*Competing interests statement:*

This research did not receive any grant from funding agencies in the public, commercial, or not-for-profit sectors. The authors declare that they have no conflict of interest.